\documentclass[twoside]{article}

\usepackage[accepted]{aistats2021}

\usepackage{microtype}
\usepackage{graphics}
\usepackage{tabularx,ragged2e,booktabs,caption}
\usepackage{esvect}
\usepackage{booktabs}
\usepackage{amsmath, amssymb}
\usepackage{bm}
\usepackage{amsthm}
\DeclareMathOperator*{\argmax}{arg\,max}
\DeclareMathOperator*{\argmin}{arg\,min}

\usepackage{wrapfig,lipsum,booktabs}
\usepackage{multirow}
\usepackage{diagbox}
\usepackage{mathtools}
\usepackage{graphicx}
\usepackage{adjustbox}
\usepackage{subfig}
\usepackage{wrapfig}
\usepackage{esvect}
\usepackage{xcolor}
\usepackage{algorithm}
\usepackage{algorithmic}
\usepackage{wrapfig}
\usepackage{mathtools}
\usepackage{bbm}
\usepackage{mathabx}
\usepackage{slashbox}
\newcommand{\defeq}{\vcentcolon=}

\usepackage{tikz}

\usepackage[colorlinks=true]{hyperref}
\hypersetup{citecolor=blue}

\usepackage{textcomp}

\usepackage{ragged2e}
\newcommand{\quotes}[1]{``#1''}
\usepackage{caption}
\usepackage{enumitem}
\usepackage{balance}

\usepackage{multicol}
\usepackage{longtable}

\usepackage{cleveref}

\DeclarePairedDelimiter\floor{\lfloor}{\rfloor}

\usepackage[square]{natbib}

\begin{document}

%

%

\twocolumn[

\aistatstitle{AutoCP: Automated Pipelines for Accurate Prediction Intervals}
\vspace{-0.25cm}
\aistatsauthor{Yao Zhang \And   William Zame \And Mihaela van der Schaar  }

\aistatsaddress{University of Cambridge \And UCLA \And University of Cambridge, UCLA\\ The Alan Turing Institute  } ]

\begin{abstract}
Successful application of machine learning models to real-world prediction problems, e.g. financial forecasting and personalized medicine, has proved to be challenging, because such settings require limiting and quantifying the uncertainty in the model predictions, i.e. providing valid and accurate prediction intervals. Conformal Prediction is a distribution-free approach to construct valid prediction intervals in finite samples. However, the prediction intervals constructed by Conformal Prediction are often (because of over-fitting, inappropriate measures of nonconformity, or other issues) overly conservative and hence inadequate for the application(s) at hand. This paper proposes an AutoML framework called Automatic Machine Learning for Conformal Prediction (AutoCP). Unlike the familiar AutoML frameworks that attempt to select the best prediction model, AutoCP constructs prediction intervals that achieve the user-specified target coverage rate while optimizing the interval length to be accurate and less conservative. We tested AutoCP on a variety of datasets and found that it significantly outperforms benchmark algorithms.
\end{abstract}

\section{Introduction}\label{sect:intro}

\begin{figure*}[t]
\centering
 \includegraphics[width=0.82\textwidth]{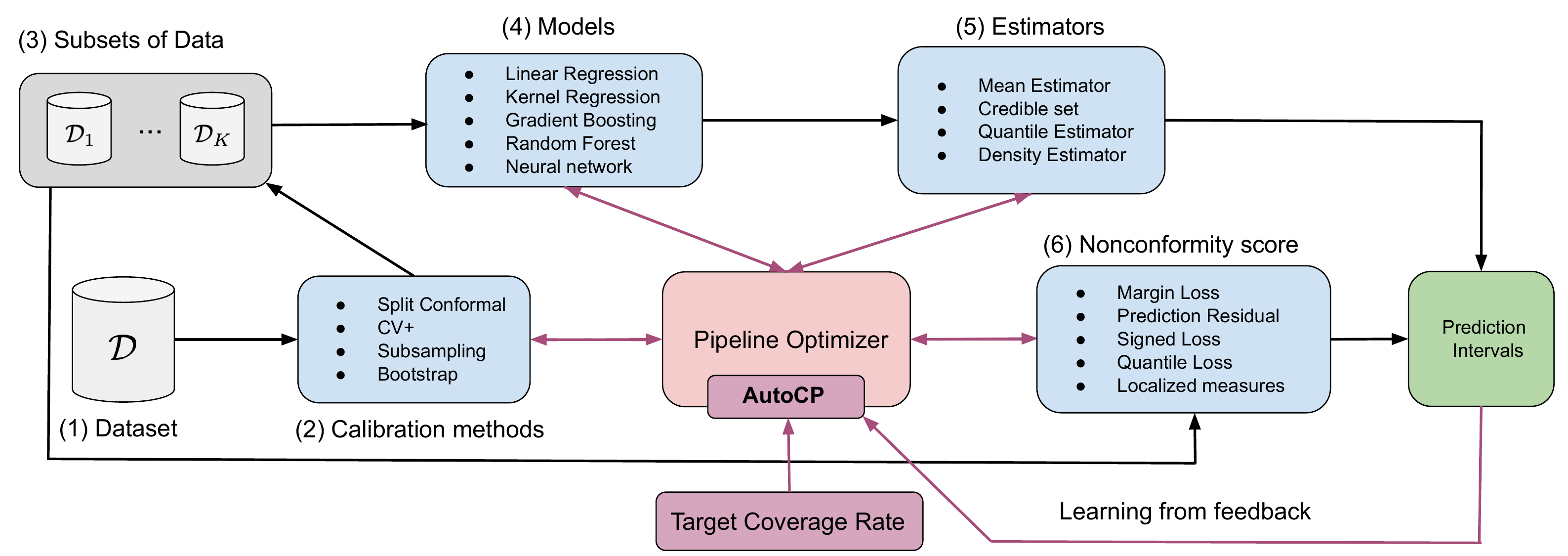}
\caption{A pictorial depiction of AutoCP. The steps of CP are enumerated in order. Implementing CP on a data set $\mathcal{D}$ is a compound decision problem, requiring the choice of a calibration method, a model, an estimator, a nonconformity score and their hyperparameters. AutoCP aims to solve this compound decision efficiently.}
\label{fig:diag1}
\end{figure*}


Machine Learning (ML) has made remarkable contributions to prediction in areas ranging from image classification to natural language processing, from search engine to advertisement. It has been less successful in areas such as finance and medicine, in large part because actionable predictions in those areas require not only predictions but also the uncertainty associated with the predictions \citep{amodei2016concrete}.  This paper proposes an AutoML framework called Automatic Machine Learning for Conformal Prediction (AutoCP), that produces prediction intervals that are tight and satisfy the Frequentist coverage guarantee. 
Without exaggerating the prediction uncertainty, these intervals cover the true labels with high probability.

AutoCP builds on the by-now familiar Automatic Machine Learning (AutoML) framework \citep{FeurerKESBH15,KotthoffTHHL17,pmlr-v80-alaa18b,zhang2019lifelong,escalante2020automl}. However, unlike these frameworks which build pipelines to optimize performance metrics such as accuracy and AUPRC, AutoCP builds a pipeline to optimize prediction intervals that satisfy the target coverage rate\footnote{The target coverage rate can be specified arbitrarily by the user.}, (e.g. construct the shortest prediction intervals with 90\% coverage). AutoCP combines conformal prediction (CP) with discriminative machine learning (ML)  models. This combination is crucial because conformal algorithms can provide the required coverage guarantee for any ML model while discriminative ML models can identify regions of high and low uncertainty. (This paper specifies a particular collection of models and algorithms, but other collections could be used.)

The pipeline optimization process of AutoCP is depicted in \Cref{fig:diag1}. Given a dataset and a target coverage rate, AutoCP constructs prediction intervals by choosing a calibration method, a model, an estimator, a nonconformity score and their hyperparameters in conformal prediction. The average length of the constructed intervals acts as the feedback for AutoCP to update its selection. To speed up the optimization process, we assume the average interval length as an unknown high-dimensional function with a factorized structure. We assume the mapping from each factor to the average interval length as a black-box function. These black-box functions are optimized jointly using Bayesian Optimization (BO) \citep{SnoekLA12} with a divide and conquer strategy. We decompose the high-dimensional problem into a set of lower-dimensional subproblems. AutoCP is tailored to sample-efficient in optimizing these black-box functions because solving each sub-problem separately takes fewer BO iterations.

We note that there would be no difficulty in marrying a given conformal algorithm with a given ML model. However, choosing by hand the conformal algorithm, the ML model and their hyperparameters to be used for a given dataset would be an extremely difficult task, requiring expertise in the various algorithms and models and specialized knowledge of the particular dataset. AutoCP automates the pipeline optimization process, thereby allowing non-experts to construct prediction intervals that satisfy the target coverage guarantee.

To demonstrate the effectiveness of AutoCP, we conduct two sets of experiments. The first set of experiments pits AutoCP against eight benchmark algorithms on eight real-world datasets from different application domains. Our results show that AutoCP achieves prediction intervals that are from 5\% to 38\% shorter than the {\em best} of the benchmark algorithms while satisfying the rigorous coverage guarantee. A separate source of gain analysis confirms the importance of the global pipeline optimization in AutoCP. The second set of experiments adapts AutoCP to the problem of estimating Conditional Average Treatment Effects, and pits AutoCP against three benchmark algorithms on two frequently-used datasets. Our results show that only AutoCP and one of the benchmark algorithms are able to achieve the required coverage, and that AutoCP achieves prediction intervals that much shorter than this competitor on both datasets. 

Following this Introduction, we first discuss Related Work of uncertainty quantification in machine learning, then lay out the Problem Formulation, describe Conformal Prediction, provide details of AutoCP, describe the Experimental Results, and present a brief Conclusion. We relegate some of the technical details and additional information about the experiments to Appendix.

\section{Related work} \label{sect:related}
Bayesian statistics offers a probabilistic framework for modelling uncertainty in the data. In the Bayesian framework, prediction uncertainty is quantified by the posterior credible sets. In practice, exact Bayesian inference is computationally prohibitive for the overparameterized deep learning models. Variational inference \citep{welling2011bayesian,gal2016dropout,hernandez2015probabilistic,maddox2019simple} approximates the true posterior distribution by an ensemble of neural networks, in the same form as the non-bayesian ad-hoc ensemble approaches \citep{lakshminarayanan2017simple}. These methods can measure prediction uncertainty by the model disagreement in the ensemble.  However, they would quantify prediction uncertainty incorrectly if the posterior distribution is approximated poorly. In general, it is impossible to construct honest credible sets (i.e. satisfying the frequentist coverage guarantee) \citep{cox1993analysis,bayarri2004interplay}. Hierarchical and Empirical Bayes methods can construct honest credible sets asymptotically under some extra assumptions on the functional parameters \citep{szabo2015frequentist,rousseau2016asymptotic}. But these assumptions are hard to verify in practice and fail to hold when we use some machine learning models such as neural network and random forest.

Conformal prediction (CP), pioneered by \cite{vovk2005algorithmic}, is a general framework for constructing prediction intervals that satisfy the frequentist coverage guarantee in finite samples. This framework is distribution free and hence applicable to any prediction model. However, naive application of conformal prediction can yield poor results; e.g. providing intervals with the same length at every point in the input space, failing to reflect that more accurate predictions are possible for some inputs than for others, especially if the data is heteroscedastic (as is often the case). A variety of localized methods have been proposed to address this problem, such as accounting for the variability of the regression function \citep{barber2019predictive},  combining conformal prediction with quantile regression \citep{romano2019conformalized,sesia2019comparison,kivaranovic2019adaptive} or reweighting samples using appropriate distance metrics \citep{guan2019conformal}. The conformal prediction literature is growing in both application and theory; we will refer to more related works in the following sections.

In our work, we use machine learning models and conformal algorithms together to achieve actionable estimates of prediction uncertainty. We develop a general and automatic framework for constructing actionable prediction intervals, in favour of practitioners' interests.

\section{Problem Formulation}\label{sect:problem}
We consider a standard regression setup of supervised learning. Let $X \in \mathcal{X}\subseteq \mathbb{R}^{d}$ denote the {\em features} and $Y\in \mathcal{Y}\subseteq\mathbb{R}$ denote  the interested label. Throughout the paper, we assume that samples are drawn exchangeably from an (unknown) distribution $P_{X,Y}$ on $\mathcal{X} \times \mathcal{Y}$.\footnote{It is standard in the literature to make the assumption that data is drawn exchangeably rather than i.i.d (which would be stronger).} We let $P_X$ denote the induced marginal on the features $X$, and $P_{Y\mid X}$ denote the conditional distribution of $Y$ given $X$. Let $\mathcal{D}=\{(X_i,Y_i)\}_{i=1}^{n}$ denote a set of $n$ training samples, and $(X,Y)$ denote a random testing sample. 
For an regression model (i.e. mean estimator) $\hat{\mu}:\mathcal{X}\rightarrow\mathcal{Y}$ fitted to $\mathcal{D}$ (so that $\hat{Y} = \hat{\mu}(X)$ is the predicted label for the testing sample $X$), we wish to provide a prediction interval $\hat{C}$ that covers the true label $Y$ with probability at least a specified {\em target coverage rate (TCR)} $1-\alpha$, where $\alpha \in (0,1)$ is a specified {\em miscoverage rate}. The interval $\hat{C}$ is constructed based on the given $\mathcal{D}$ and  $\alpha$. We view $\mathcal{D}$ and $\alpha$ as fixed throughout the exercise and so suppress them in the notation of $\hat{C}$. In what follows, we write $|\hat{C}|$ for the length of the interval $\hat{C}$. In order that our intervals are actionable in practice, we require two things:

\textbf{Frequentist Coverage.} The prediction intervals should be marginally valid, i.e. satisfying the marginal coverage guarantee such that
\begin{equation}\label{equ:marginal_validity}
    \mathbb{P} \big\{ Y \in \hat{C}(X) \big\} \geq 1-\alpha,
\end{equation}
where the probability is taken with respect to $P_{X,Y}$. \Cref{equ:marginal_validity} means that $\hat{C}$ covers the true testing label $Y$ with probability at least $1-\alpha$, on average over a random draw of the training and testing samples $\mathcal{D}\cup\{(X,Y)\}$  from the distribution $P_{X,Y}$. Marginal validity is weaker than conditional validity which requires $\hat{C}$ to satisfy the conditional coverage guarantee,
\[
\mathbb{P} \big\{ Y \in \hat{C}(X) \mid X = x \big\} \geq 1-\alpha, \forall x\in \mathcal{X} \ \text{s.t.}\  P_X(x)>0.
\] 
This means that for every possible value $x$ of the test sample $X$, the interval $\hat{C}$ covers the true label $Y$ with probability at least $1-\alpha$. \cite{lei2015distribution,vovk2012conditional} prove that condition validity is impossible to attain in any meaningful sense; the conditionally valid prediction intervals $\hat{C}$ have infinite expected length for $x$ in the support of $P_X$. Marginally valid intervals only achieve the TCR over $\mathcal{X}$ on average, but it may fail to achieve the TCR for some specific regions of $\mathcal{X}$. We note that it is possible to divide the space $\mathcal{X}$ into subsets and construct $\hat{C}$ using the samples in each subset so that the TCR is achieved in each subset \citep{lei2015distribution,lee2020robust,barber2019limits}. Dividing samples into subsets is likely to inflate the interval length due to the sample scarcity in each subset. Throughout the paper, we want our prediction intervals to achieve the marginal coverage guarantee in \Cref{equ:marginal_validity}, as the minimum requirement for our prediction intervals to be actionable.

\textbf{Discrimination.} On average, prediction intervals are wider for testing samples with large variance.  That is, for two different values $x$, $x'\in\mathcal{X}$, we have
\begin{equation*}
{\rm Var}\big(Y|X=x\big)\leq {\rm Var}\big(Y|X=x'\big)\Rightarrow |\hat{C}(x)| \leq |\hat{C}(x')| 
\end{equation*}
where the variance ${\rm Var}(\cdot)$ is taken with respect to the underlying data distribution $P_{Y|X}$. Discrimination does not come with a guarantee. But it provides a guess of the high and low variance region in $\mathcal{X}$ so that we can adapt the interval length $|\hat{C}|$ locally.

Both requirements are crucial for prediction intervals to be actionable. If the intervals fail to achieve the TCR, a user can not completely trust them in practice. On the other hand, conformal algorithms can create wide intervals to satisfy the TCR but exaggerate the prediction uncertainty; the user would distrust the model predictions even when they are accurate. Conformal algorithms can construct marginally valid prediction intervals under the mild assumption (weaker than i.i.d.) that the training and testing samples are drawn exchangeably from the same data distribution $P_{X,Y}$. Therefore, the prediction intervals constructed by the standard conformal algorithms do not serve to solve the problem of out-of-distribution detection. Nevertheless, prediction models still make errors on in-distribution samples; prediction intervals provide bounds on the magnitude of these errors. These bounds are important to a potential user (e.g. a clinician making treatment decisions), and tighter bounds mean the potential user can be more certain about the model predictions. Constructing tight and marginally valid prediction intervals is the purpose of AutoCP.

\section{Conformal Prediction}\label{sect:nested}
In this section, we summarize a variety of conformal algorithms that can be used in our AutoML framework. 

\subsection{Mean estimators}
We start by introducing a simple method called Split Conformal Prediction (SCP) \citep{lei2018distribution}. In SCP we first split the training set into two equal-size subsets $\mathcal{I}_1$ and $\mathcal{I}_2$. Let $\hat{\mu}_{\mathcal{I}_1}(\cdot)$ denote the mean estimator fitted to $\mathcal{I}_1$. 
We define the nonconformity score as the residual
\begin{equation}\label{equ:rxy_split}
    r(X,Y)  = |\hat{\mu}_{\mathcal{I}_1}(X)-Y|.
\end{equation}
We collect the residuals of $\hat{\mu}_{\mathcal{I}_1}(\cdot)$  on the validation samples in $\mathcal{I}_2$ and construct the set of scores
\begin{equation}\label{equ:ri2}
   R_{\mathcal{I}_2}=\{r_i = r(X_i,Y_i):(X_i,Y_i)\in \mathcal{I}_2\}.
\end{equation}
The prediction interval of SCP takes the form
\[
 \hat{C}_{t}(X) = \big[\hat{\mu}_{\mathcal{I}_1}(X)-t,    \hat{\mu}_{\mathcal{I}_1}(X)+t\big].
\]
Taking $t$ as $\hat{Q}_{1-\alpha}^{\mathcal{I}_2}\defeq(1-\alpha)(1+1/|\mathcal{I}_2|)$-th quantile of $R_{\mathcal{I}_2}$, i.e. the $\floor{(|\mathcal{I}_2|+1)\alpha}$-th largest residual in $\mathcal{I}_2$, the SCP prediction interval is given as
\begin{equation}\label{equ:interval_scf}
    \begin{split}
        \hat{C}_{\text{split}}(X)=\big[\hat{\mu}_{\mathcal{I}_1}(X)-\hat{Q}_{1-\alpha}^{\mathcal{I}_2},    \hat{\mu}_{\mathcal{I}_1}(X)+\hat{Q}_{1-\alpha}^{\mathcal{I}_2}\big].
    \end{split}
\end{equation}
The following is the intuition of why $\hat{C}_{\text{split}}$ is marginally valid. Exchangeability implies that the joint distribution of the validation and testing residuals is invariant under any permutation of the residuals. The rank of the testing residual $R(X,Y)$ is uniformly distributed in $\{1,\dotsc,|\mathcal{I}_2|+1\}$; $R(X,Y)$ is smaller than the $\floor{(|\mathcal{I}_2|+1)\alpha}$-th largest residual in $\mathcal{I}_2$ with probability at least $1-\alpha$. Therefore, $\hat{C}_{\text{split}}(X)= \hat{C}_{t}(X) $ with $t=\hat{Q}_{1-\alpha}^{\mathcal{I}_2}$ satisfies the marginal coverage guarantee in \Cref{equ:marginal_validity}. We note $\hat{Q}_{1-\alpha}^{\mathcal{I}_2}$ is the smallest $t$ that we can choose to achieve the guarantee.

From SCP, we can see the process of CP consists of four steps: (1) Fitting a estimator $\hat{\mu}_{\mathcal{I}_1}(\cdot)$ to the training set $\mathcal{I}_1$; (2) Define the nonconformity score $r(X,Y)$; (3) Construct the set of scores $R_{\mathcal{I}_2}$ on the validation set $\mathcal{I}_2$; (4) Construct the prediction interval with the $\floor{(|\mathcal{I}_2|+1)\alpha}$-th largest score in $R_{\mathcal{I}_2}$. Note that the length of $\hat{C}_{\text{split}}$ is constant over the input space and hence non-discriminative. This problem motivates to generalize SCP in two different directions: (1) interval estimator methods, and (2) calibration methods.

\subsection{Interval estimators}\label{sect:interval}
When the data is heteroscedastic, we can construct discriminative prediction intervals by replacing the mean estimator $\mu_{\mathcal{I}_1}(\cdot)$ in SCP with interval estimators such as Bayesian credible sets and quantile functions.

The method Locally Weighted Split Conformal \citep{lei2018distribution} learns both the mean estimator $\hat{\mu}_{\mathcal{I}_1}(\cdot)$ and mean absolute deviation (MAD) estimator $\hat{\sigma}_{\mathcal{I}_1}(\cdot)$ from the training split $\mathcal{I}_1$. The MAD estimator $\hat{\sigma}_{\mathcal{I}_1}(\cdot)$ is obtained by fitting a separate regression model to the fitting residuals of $\hat{\mu}_{\mathcal{I}_1}(\cdot)$ on $\mathcal{I}_1$. For Bayesian models, the mean and MAD estimator can be directly derived from the posterior distribution. After fitting estimators $\hat{\mu}_{\mathcal{I}_1}(\cdot)$ and  $\hat{\sigma}_{\mathcal{I}_1}(\cdot)$ to $\mathcal{I}_1$, we define the nonconformity score as the normalized residual
\begin{equation}\label{equ:rxy}
    r(X,Y)  = \hat{\sigma}_{\mathcal{I}_1}^{-1}(X)|\hat{\mu}_{\mathcal{I}_1}(X)-Y|.
\end{equation}
We redefine $R_{\mathcal{I}_2}$ in \Cref{equ:ri2} with the new $r(X,Y)$ in \Cref{equ:rxy}. Taking $t$ as the $\hat{Q}_{1-\alpha}^{\mathcal{I}_2} \defeq \floor{(|\mathcal{I}_2|+1)\alpha}$-th largest score in $R_{\mathcal{I}_2}$, we obtain the prediction interval
\begin{equation}\label{equ:interval_local}
\begin{split}
\hat{C}_{\text{loc}}(X) = & \big[\hat{\mu}_{\mathcal{I}_1}(X)-\hat{Q}_{1-\alpha}^{\mathcal{I}_2}\hat{\sigma}_{\mathcal{I}_1}(X),\\
&\  \hat{\mu}_{\mathcal{I}_1}(X)+\hat{Q}_{1-\alpha}^{\mathcal{I}_2}\hat{\sigma}_{\mathcal{I}_1}(X)\big] .
\end{split}
\end{equation}
Conformal Quantile Regression \citep{romano2019conformalized} is another variant of SCP based on quantile regression. Let $\hat{q}_{\alpha/2,\mathcal{I}_1}(\cdot)$ and $\hat{q}_{1-\alpha/2,\mathcal{I}_1}(\cdot)$ be two conditional quantile functions fitted to $\mathcal{I}_1$. The nonconformity score is the out-of-quantiles residual
\begin{equation}\label{equ:rxy_quantiles}
    r(X,Y)=\max \{\hat{q}_{\alpha/2,\mathcal{I}_1}(X)-Y, Y-\hat{q}_{1-\alpha/2,\mathcal{I}_1}(X) \}.
\end{equation}
We define $R_{\mathcal{I}_2}$ in \Cref{equ:ri2} with the new nonconformity score in \Cref{equ:rxy_quantiles}.
Taking $t$ as the $\hat{Q}_{1-\alpha}^{\mathcal{I}_2} \defeq\floor{(|\mathcal{I}_2|+1)\alpha}$-the largest score in $R_{\mathcal{I}_2}$, we obtain the prediction interval
\begin{equation}\label{equ:interval_cqr}
\begin{split}
	\hat{C}_{\text{cqr}}(X) = & [\hat{q}_{\alpha/2,\mathcal{I}_1}(X)-\hat{Q}_{1-\alpha}^{\mathcal{I}_2},    \\
&	\ \hat{q}_{1-\alpha/2,\mathcal{I}_1}(X)+\hat{Q}_{1-\alpha}^{\mathcal{I}_2}].
\end{split}
\end{equation}
After some algebra, we can see that the prediction intervals
$\hat{C}_{\text{split}}(X)$, $\hat{C}_{\text{loc}}(X)$ and $\hat{C}_{\text{cqr}}(X)$, in \Cref{equ:interval_scf,equ:interval_local,equ:interval_cqr} respectively, are given in the same form as 
\[
\Big\{y: r(X,y) \leq \hat{Q}_{1-\alpha}^{\mathcal{I}_2}  \Big\}.
\]
The main differences in these three algorithms are the used estimators and nonconformity scores in \Cref{equ:rxy_split,equ:rxy,equ:rxy_quantiles}.  
The lengths of $\hat{C}_{\text{cqr}}(X)$ and $\hat{C}_{\text{loc}}(X)$ vary over the space $\mathcal{X}$ while the length of $\hat{C}_{\text{split}}(X)$ is constant. For example, $\hat{C}_{\text{cqr}}(X)$ is wide, i.e. the gap between the upper quantile $\hat{q}_{\alpha/2,\mathcal{I}_1}(X)$ and the lower quantile $\hat{q}_{1-\alpha/2,\mathcal{I}_1}(X)$ is large, 
indicating large uncertainty of the label $Y$. Using an interval estimator to construct discriminative prediction intervals is a popular idea in the literature. In Table {\color{red}4} of Appendix {\color{red}A}, we summarize other conformal algorithms using the same idea but with different estimators and prediction intervals. The process of constructing prediction intervals is similar in these conformal algorithms. We note that an interval estimator does not issue a prediction of the mean of $Y$. In practice, we can train another mean estimator to predict $Y$, and only use the prediction intervals returned by the conformal algorithm to quantify the uncertainty of the label $Y$.

\subsection{Calibration methods}\label{sect:calibration}
In SCP and its extensions, the training samples are spitted into two equal-size subsets, $\mathcal{I}_1$ and $\mathcal{I}_2$. One could consider other calibration methods, such as leave-one-out \citep{barber2019predictive}, $K$-fold cross-validation \citep{vovk2015cross} and bootstrap method \citep{kuchibhotla2019nested,kim2020predictive}. These variants enable $\hat{\mu}(\cdot)$ to be trained on more samples and have smaller nonconformity score on the validation samples. The prediction intervals obtained by $K$-fold cross-validation and Bootstrap are given in Equations ({\color{red}12}) and ({\color{red}13}) of Appendix {\color{red}B}. Like in conventional model selection, K-fold splitting tends to overestimate the true prediction error while Bootstrap tends to underestimate the true prediction error \citep{hastie2009elements}. Both methods can achieve the coverage guarantee, as summarized in Table {\color{red}5} of Appendix {\color{red}B}. AutoCP aims to select the best out of the two calibration methods and optimize the number of folds for constructing tight prediction intervals.

\section{AutoCP}\label{sect:AutoCP}

In the last section, we demonstrate that a user can design a conformal algorithm with a variety of estimators and calibration methods. The user also needs to choose a machine learning model and its hyperparameters for parametrizing the chosen estimator. We view all these choices together as a pipeline optimization problem for constructing prediction intervals, as illustrated earlier in \Cref{fig:diag1}. We now introduce our framework AutoCP which solves the pipeline optimization problem.

Fix a set $[M] = \{1, \ldots, M\}$ of models; for each model, we will choose hyperparameters $\Lambda_m$; write ${\mathcal A}_{[M]}=\bigtimes_{m\in[M]}\mathcal{A}_m$ for the set of pairs consisting of a model $m \in [M]$ and a choice set of hyperparameters.  Fix a set 
${\mathcal A}_e$ of estimators and a set  ${\mathcal A}_c$ of calibration methods. A pipeline is  4-tuple 
$(m, \Lambda_m, \Lambda_e, \Lambda_c)$ consisting of a model, a set of  hyperparameters for that model, an estimator and a calibration method; the space of all possible pipelines is $\mathcal{P} = \mathcal{A}_{[M]} \times\mathcal{A}_e\times\mathcal{A}_c$. An example pipeline might be $\{$Random Forest, 1000 stumps using 4 features, Quantile estimator, 5-fold cross-validation$\}$. Each model would have many possible sets of hyperparameters. The space $\mathcal{P}$ of pipelines would be high-dimensional, even if we restrict our attention to a few models, estimators and calibration methods. The goal of AutoCP is to identify the pipeline $\tilde{\Lambda}^{*}\in\mathcal{P}$ for a given dataset that yields the minimum average interval length on the validation samples:
\begin{equation}\label{equ:blackbox}
\begin{split}
\tilde{\Lambda}^{*} &\in \argmin_{\tilde{\Lambda}\in \mathcal{P}} \tfrac{1}{J}\sum_{j=1}^{J} \tfrac{1}{|\mathcal{D}_{\text{val}}^{(j)}|}\sum_{(X_i,Y_i)\in \mathcal{D}_{\text{val}}^{(j)}} |\hat{C}_{\tilde{\Lambda}} (X_i)| \\ 
\end{split}
\end{equation}
where $J$ is the chosen number of folds in cross-validation, $\mathcal{D}_{\text{val}}^{(j)}$ is the validation set of the $j$th split, and $|\hat{C}_{\tilde{\Lambda}} (X_i)|$ is the interval length for the validation sample $X_i$. We note that the marginal coverage guarantee in \Cref{equ:marginal_validity} is satisfied by applying any conformal algorithm. The objective in (\ref{equ:blackbox}) has no analytic form. We now take a divide and conquer strategy to solve this optimization problem. For each model $m\in[M]$, we assume the average interval length of the associated pipeline $\tilde{\Lambda}_m=[\Lambda_m,\Lambda_e,\Lambda_c]$,
\[
 \tfrac{1}{J}\sum_{j=1}^{J} \tfrac{1}{|\mathcal{D}_{\text{valid}}^{(j)}|}\sum_{(X_i,Y_i)\in \mathcal{D}_{\text{valid}}^{(j)}} |\hat{C}_{\tilde{\Lambda}_m}(X_i)|,
\]
is a noisy version of a black-box function $\tilde{f}_m:\mathcal{P}_m\rightarrow \mathbb{R}$. Let $\mathcal{P}_m = \mathcal{A}_{m} \times\mathcal{A}_e\times\mathcal{A}_c$ denote the space of all possible pipelines with model $m$. Next we use Bayesian Optimization (BO) \citep{SnoekLA12} to find the minimum of each $\tilde{f}_m$, i.e. the optimal pipeline $\tilde{\Lambda}_m^{*}\in\mathcal{P}_m$ with model $m$. The BO algorithm specifies a Gaussian process (GP) prior on each $\tilde{f}_m(\cdot)$,
\begin{equation}\label{equ:gp}
	\tilde{f}_m\sim \mathcal{GP}_m (\tilde{\mu}_m(\tilde{\Lambda}_m),\tilde{k}_m(\tilde{\Lambda}_m,\tilde{\Lambda}_m') )
\end{equation}
where $\tilde{\mu}_m(\tilde{\Lambda}_m)$ is the mean function and $\tilde{k}_m(\tilde{\Lambda}_m,\tilde{\Lambda}_m')$ is the covariance kernel that measures the similarity between two different pipelines with model $m$. Our BO algorithm is a simple three-step iterative procedure to find the solution of \Cref{equ:blackbox}:
\begin{enumerate}
    \item Fit a GP model to the pipeline evaluation data of model $m$ and update the acquisition function $\tilde{a}_{m}$ of model $m$, for every $m\in [M]$;
    \item Search for a optimizer $\tilde{\Lambda}_m^{*}$ of $\tilde{f}_m$ by optimizing the acquisition function $\tilde{a}_{m}$ and select the model $m^{*}\in \argmax_m \tilde{a}_{m}(\tilde{\Lambda}_{m}^{*})$ to evaluate next;
   \item Evaluate the pipeline $\tilde{\Lambda}_{m^{*}}^{*}$ and add the pipeline performance to the pipeline evaluation data for the next fitting of the GP models in Step 1.
\end{enumerate}
The acquisition function $\tilde{a}_{m}$ \citep{kushner1964new,srinivas2012information,mockus2012bayesian} is computed by the exact GP posterior distribution. Optimizing $\tilde{a}_{m}$ in Step 2 balances the exploration and exploitation trade-off in finding the optimizer $\tilde{\Lambda}_m^{*}$ of $\tilde{f}_m$. Re-fitting the GP models (taking a few seconds for each $m$) in Step 1 is much cheaper the pipeline evaluation (Step 3) in most cases. Therefore, we want our BO algorithm to be sample efficient i.e. finding the optimal pipeline with as few evaluations as possible.


\textbf{Sample Efficiency.} Without involving any redundant dimensions in the kernel functions, our divide and conquer strategy
(modelling the pipeline performance under each model $m$ with a separate GP model) is more sample-efficient than modelling them jointly. In the joint modelling strategy, we would let $\tilde{f} = \sum_{m=1}^{M}\mathbbm{1}(v=m)\tilde{f}_m$ denote the joint black-box function  where $v$ indicates the chosen model. Let $k_{\text{joint}}(\cdot,\cdot)$ denotes the kernel function in the joint GP model for the function $\tilde{f}$ whose input $\Lambda\in\mathcal{P} = \bigtimes_{m\in[M]}\mathcal{A}_m \times\mathcal{A}_e\times\mathcal{A}_c$ is high-dimensional, including a one-hot vector indicating the chosen model and the hyperparameters of all the models. When comparing two pipeline $\Lambda$ and $\Lambda'$ with the same model $m$, $k_{\text{joint}}(\cdot,\cdot)$ takes into account the irrelevant dimensions in other $\mathcal{A}_{m'}$, $m'\neq m$, when the similarity between $\bar{\Lambda}$ and $\bar{\Lambda}'$ only depends on the input variables in $\mathcal{A}_m\times\mathcal{A}_e\times\mathcal{A}_c$. 

Even in our strategy, the model pipeline space $\mathcal{P}_m$  can be high-dimensional. High dimensionality \citep{kandasamy2015high,gyorfi2006distribution} can render the GP-based BO algorithm infeasible. We deal with this problem using the idea of sparse additive kernel decomposition \citep{kandasamy2015high,wang2017batched}. We assume the underlying structure in $\tilde{k}_{m}(\tilde{\Lambda}_m,\tilde{\Lambda}_m')$ that relates the hyperparameters of the model $m$, the choice of estimator and calibration method can be expressed via the following sparse additive kernel decomposition:
\begin{equation*}
	\tilde{k}_m(\tilde{\Lambda}_m,\tilde{\Lambda}_m') = k_{m}(\Lambda_{m},\Lambda_{m}^{'}) + k_{e}(\Lambda_{e},\Lambda_{e}^{'}) +k_{c}(\Lambda_{c},\Lambda_{c}^{'})
\end{equation*}
where $\Lambda_{m}\in \mathcal{A}_m$, $\Lambda_{e}\in \mathcal{A}_e$, and $\Lambda_{c}\in \mathcal{A}_c$. We let the kernel functions $\tilde{k}_m(\tilde{\Lambda}_m,\tilde{\Lambda}_m')$, $m\in[M]$, share the same \quotes{estimator} kernel $k_{e}(\Lambda_{e},\Lambda_{e}')$ and \quotes{calibration} kernel $k_{c}(\Lambda_{c},\Lambda_{c}^{'})$. The hyperparameters of $ k_{e}(\cdot,\cdot)$ and $k_{c}(\cdot,\cdot)$ can be learned more efficiently by maximizing the sum of marginal likelihood functions \citep{williams2006gaussian} over all the functions $\tilde{f}_m$, $m\in[M]$. The decomposition breaks down the function $\tilde{f}_m(\Lambda)$ as
\begin{equation}\label{equ:additive_function}
	\tilde{f}_m(\tilde{\Lambda}_m) = f_m(\Lambda_m) + f(\Lambda_e) + f(\Lambda_c) 	
\end{equation}
The sparse additive structure in (\ref{equ:additive_function}) gives a statistically efficient BO algorithm. That is, if a function is $\gamma$-smooth, the additive kernels reduce sample complexity from
$O(n^{\frac{-\gamma}{2\gamma+D}})$ to $O(n^{\frac{-\gamma}{2\gamma+D_s}})$, where $D_s$ is the maximum number of dimensions in any subspace \citep{yang2015minimax}. Appendix {\color{red}A.3} demonstrates that our BO algorithm converges faster and finds a better minimum than the joint modelling strategy on the datasets in \Cref{tab:length_coverage}. This shows the benefit of the components in our BO algorithm: (1) the divide and conquer strategy; (2) sparse additive kernel decomposition; (3) the joint marginal likelihood optimization to learn the shared kernels $k_{e}(\Lambda_{e},\Lambda_{e}^{'})$ and $k_{c}(\Lambda_{c},\Lambda_{c}^{'})$.

It is worth noting that we can also modify the acquisition function to reduce the pipeline evaluation time. The pipeline evaluation time increases with an increasing number of folds in calibration. We can model the computational time $\tilde{g}_m(\cdot)$ using a separate GP model, and acquire a pipeline (Step 2 of our BO algorithm) with the acquisition function $\tilde{a}_{m}(\tilde{\Lambda}_{m})/\tilde{g}_m(\tilde{\Lambda}_m)$ \citep{klein2016fast}. Then BO will prioritize the pipeline with low complexity, and only acquire the expensive pipelines if there is a clear trend of performance gain from using them. The implementation details of AutoCP are given in the next section and Appendix {\color{red}A.2}. 

\section{Experiments}\label{sect:experiments}

\begin{table*}[t]
\centering
\begin{small}
\caption{Comparison of the average coverage rate and interval length (AutoCP v.s. the best and worst benchmark for each of the eight datasets): AutoCP yields large improvements of interval length while the coverage rates are very similar among the algorithms. AutoCP outperforms the best model, which shows the gain of selecting good estimators and calibration methods jointly.}
\label{tab:length_coverage}
\setlength\tabcolsep{1pt}
\begin{adjustbox}{width=2\columnwidth,center}
\begin{tabular}{c|cc|cc|cc|cc}	
\toprule
\multirow{2}{*}{Mean$\pm$Std}  & \multicolumn{2}{c|}{AutoCP}  &  \multicolumn{2}{c|}{Best Benchmark} &  \multicolumn{2}{c|}{Worst Benchmark}     &  \multicolumn{2}{c}{Length Decrease ($\%$) } 
  \\
  & Avg. Coverage ($\%$) & Avg. Length & Avg. Coverage ($\%$) & Avg. Length & Avg. Coverage ($\%$) & Avg. Length &v.s. Best  &  v.s. Worst   \\
\midrule 
Community   & $90.97\pm 1.83 $ & $1.49\pm 0.03$ &$90.15\pm 2.06 $ & $1.57\pm0.09$ &$90.15\pm 2.06$ &$2.01\pm 0.19$ & $ 5.36 $  & $ 25.87$\\  
\midrule
Concrete   & $90.90\pm 2.95$ & $0.44\pm 0.02$ &$90.07\pm 2.15$ &$0.47\pm 0.03$& $90.01\pm 2.38$ & $0.92\pm 0.06$ & $ 6.38 $ &  $ 52.17$\\
\midrule
Boston  & $90.42\pm 2.75$ & $ 0.42\pm 0.03$ & $89.95\pm 3.21$ &$0.47\pm 0.04$ & $89.22\pm 4.18$ &$0.62\pm 0.06 $ & $ 10.63 $ &  $ 32.26$\\
\midrule
Energy   & $91.36\pm 2.15$ & $0.08\pm 0.01$ &$91.40\pm 3.64$ &$0.09\pm 0.01$& $90.45\pm 2.21$ & $0.54\pm 0.22$ & $ 11.11 $ &  $ 85.19$\\
\midrule
MEPS 19  & $90.41\pm  0.71$  &  $1.73\pm 0.12$ & $90.14 \pm 0.78 $ &$2.36\pm0.18$& $90.06\pm 0.58 $ &$4.67\pm 0.19$ & $ 36.41 $ &  $ 62.96$\\
\midrule
MEPS 20   & $90.09\pm 0.52$  & $1.85\pm 0.13$ & $89.84\pm 0.54$  &$2.45\pm0.19$& $89.99\pm 0.65 $ & $4.63\pm 0.14 $  & $ 24.48 $ &  $ 60.04$\\
\midrule
MEPS 21 & $90.21\pm 0.66$ & $1.80\pm 0.11$ & $90.03\pm 0.57$ &$2.37\pm0.17$& $90.01\pm 0.53 $ & $4.75\pm 0.19 $ & $ 24.05 $ &  $ 63.17$\\
\midrule
Wine & $91.40 \pm 1.70$ &  $0.22\pm 0.02$ & $90.34\pm 2.49$  &$0.36 \pm 0.02$& $91.94\pm 3.64$ &$0.42\pm 0.07$ & $ 38.89 $ &   $ 47.61$\\
\bottomrule
\end{tabular}
\end{adjustbox}
\vspace{-3pt}
\centering
\end{small}
\end{table*}

We conducted two sets of experiments.  The first set of experiments employs eight regression datasets\footnote{The first five datasets are UCI datasets \citep{Dua:2019} and the MEPS datasets are in: https://meps.ahrq.gov/mepsweb.}: Community and Crimes (Community), Boston Housing (Boston), concrete compressive strength (Concrete), Red wine quality (Wine), Energy Efficiency (Energy), and three Medical Expenditure Panel Surveys (MEPS 19, MEPS 20, MEPS 21). We fix the miscoverage rate $\alpha = 0.1$, so the target coverage rate is 90\%.\footnote{The coverage rates in the experiments differ slightly from 90\% because we are dealing with finite samples.}  For each dataset, we construct 20 random splits, with $80\%$ of the samples used for training and the remaining $20\%$ for testing, and we average the performance over these 20 splits. Conformal algorithms \citep{romano2020classification,sadinle2019least,guan2019prediction,cauchois2020knowing} and AutoCP extend seamlessly to classification problems. Classification requires a different type of nonconformity scores, e.g. distance to the nearest neighbours. The prediction intervals may contain more than one class to cover the discrete label $Y$.  Here, we only focus on regression datasets because the interval length is a more intuitive measure of discrimination; it is more challenging because the interval needs to cover a continuous region of $Y$; it is a more commonly used metric in the literature.

\textbf{Benchmarks}. We compare our autoML framework AutoCP with a total of eight benchmark  algorithms which have valid finite sample coverage guarantees. The underlying models we consider are Ridge Regression, Random Forest, and Neural Network. Neural Network here is a standard Multiple Layer Perceptron. The benchmarks consist of (a) the original version of SCP for each of these models, denoted SCP-Ridge, SCP-RF, and SCP-NN; (b) the locally adaptive conformal prediction version of SCP \citep{lei2018distribution} for each of these models, denoted SCP-Ridge-Local, SCP-RF-Local, and SCP-NN-Local; (c)\footnote{These methods subsume the standard conformal variants of random forests \citep{johansson2014regression} and neural networks \citep{papadopoulos2011reliable}.} Conformalized Quantile Regression (CQR) \citep{romano2019conformalized}, using Neural Network and Random Forest as underlying models, denoted CQR-NN and CQR-RF. In our implementation of AutoCP, we use Ridge Regression, Random Forest, Neural Network as underlying models, the first five estimators listed in Table {\color{red}4} in Appendix {\color{red}A}, and the calibration methods shown Table 5 of Appendix {\color{red}B}. (Further details of the implementation of the benchmark algorithms and of AutoCP are provided in Appendix {\color{red}A}.)

\begin{figure}[H]
\centering
 \includegraphics[width=0.4\textwidth]{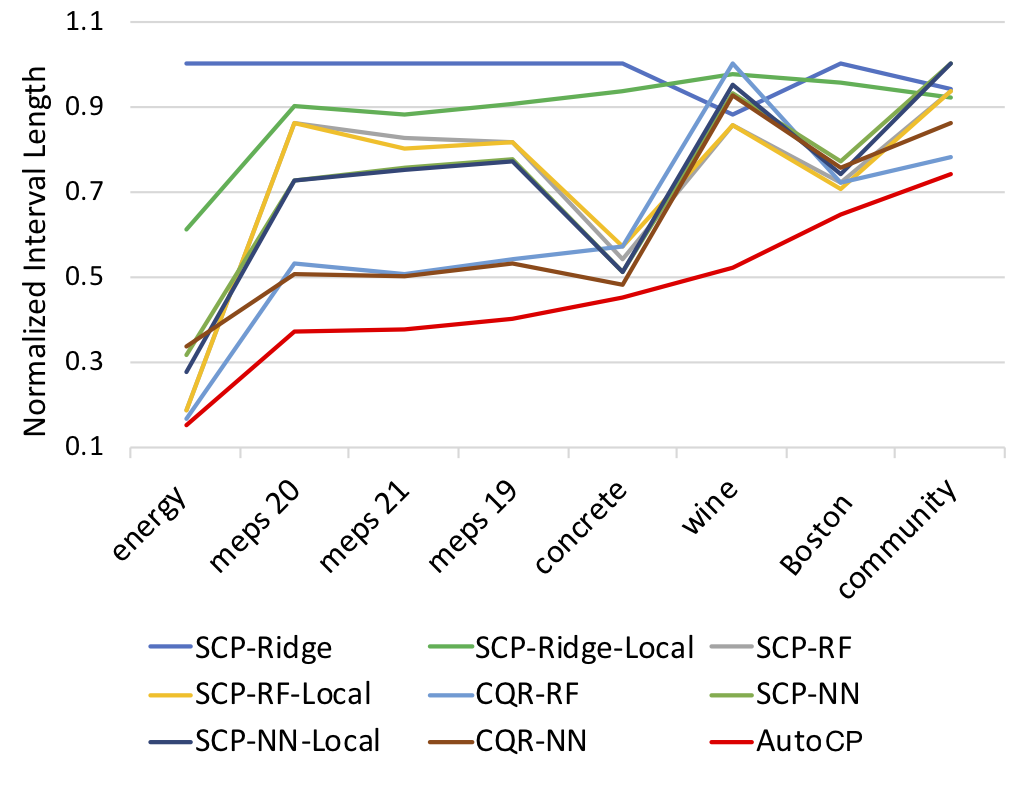}
\caption{Normalized Interval Lengths}
\label{fig:gain_data90}
\vspace{-5pt}
\end{figure}

\textbf{Performance results.} For AutoCP and each of the eight benchmarks, we compute the average coverage rate and interval length for each dataset.  The relative performance of each algorithm on each dataset is displayed in Figure \ref{fig:gain_data90}: for each dataset we normalize by dividing by the average interval length for the {\em worst-performing algorithm} (so scores are all in the interval $(0,1]$.) (To aid in legibility, we ordered the presentation of the datasets according to the relative performance of AutoCP.) As is easily seen, AutoCP displays the best performance on {\em every} dataset. Table \ref{tab:length_coverage} compares the performance of AutoCP against the Best and Worst Benchmarks on each dataset. As can be seen, all nine algorithms achieve the target coverage rate of 90\% (or very close to it) on all datasets, but the Length (of  prediction intervals) varies widely across datasets and across algorithms.  The last two columns show the percentage improvement in Length (of  prediction intervals) that AutoCP achieves over the Best and Worst Benchmarks.  We highlight that the improvement of AutoCP over the Best Benchmark ranges from 5.36\% on the Community dataset to 38.89\% on the Wine dataset.

As noted in the Introduction, it is not difficult to marry a conformal algorithm and a machine learning model, but the ``right'' algorithm and model to marry will depend on the particular dataset.  This can be seen in Figure \ref{fig:gain_data90}: CQR-NN is the best-performing benchmark (marriage of an algorithm and a model) on the MEPS 19, 20, 21 and Concrete datasets (and still AutoCP outperforms it by 36.41\%, 24.48\%, 24.05\% and 6.38\%, respectively), but CQR-NN is in the middle of the pack on the Energy, Wine, Boston and Community datasets (where AutoCP outperforms it by 55.56\%, 43.59\%, 14.28\% and 13.87\%, respectively).

\textbf{Source of gain}. To better understand where and how the gains achieved by AutoCP arise, we consider the effects of fixing a particular component and then optimizing the others.  We consider two cases: In  \quotes{Model+Cal}, we fix the estimator to be conformal quantile regression and optimize the selection of prediction models and calibration method; in \quotes{Estimator+Cal}, we fix the model to be a neural network and optimize the selection of estimator and calibration method.  We report the results in Table \ref{tab:gain}. Both of the restricted variants are out-performed by the full version of AutoCP on all the datasets. The improvement of AutoCP over either restricted variants is quite significant.  This supports the view that we should optimize the selection of {\em all} components of the pipeline, and not just of a portion.

\begin{table}[H]
\centering
\begin{small}
\caption{Source of gain}
\setlength\tabcolsep{1pt}
\begin{adjustbox}{width=0.95\columnwidth,center}
\begin{tabular}{c|c|c|c}	
\toprule
\multirow{2}{*}{Datasets}  & \multicolumn{3}{c}{Avg. Length} \\
   & Model + Cal & Estimator + Cal & AutoCP\\
\midrule
Boston  &  $0.47\pm 0.02$ &  $0.45 \pm 0.02 $ & $\bm{0.42\pm 0.03}$ \\
\midrule
Energy   & $0.16\pm 0.02$  &  $0.19 \pm 0.01$ & $\bm{0.08\pm 0.01}$ \\
\midrule
Wine & $0.28\pm 0.03 $ & $0.34\pm 0.02$  & $\bm{0.22\pm 0.02}$  \\
\midrule
MEPS 19   &  $2.16\pm 0.14$ & $2.36 \pm 0.14$ & $\bm{1.73\pm 0.12}$ \\
\midrule
MEPS 20    &  $2.15\pm 0.14$ &  $2.39 \pm 0.13$ & $\bm{1.85\pm 0.13}$ \\
\midrule
MEPS 21  &  $2.12 \pm 0.13$ &  $2.38 \pm 0.15$ & $\bm{1.80\pm 0.11}$ \\
\midrule
Concrete   & $0.45\pm 0.03$  & $0.46\pm 0.02$& $\bm{0.44\pm 0.02}$  \\
\midrule
Community  &  $1.69\pm 0.04$ & $1.68\pm 0.02$ & $\bm{1.49\pm 0.03}$ \\
\bottomrule
\end{tabular}
\end{adjustbox}
\label{tab:gain}
\vspace{-5pt}
\end{small}
\end{table}


For the second set of experiments, we turn to apply conformal algorithms to develop reliable prediction intervals for Conditional Average Treatment Effects (CATE). The application of conformal algorithms on CATE estimation is recently demonstrated in \citep{lee2020robust,kivaranovic2020conformal,lei2020conformal}. If the responses are $Y(0)$ (untreated) and $Y(1)$ (treated) then the CATE is $\mathbb{E}[Y(1)-Y(0) | X=x]$.\footnote{In actual data, either the patient was treated or not-treated but not both -- so only one of $Y(0), Y(1)$ is observed.} State-of-the-art methods for  CATE estimation rely either on Bayesian credible sets \citep{chipman2010bart,alaa2017bayesian,zhang2020learning} or on (tree-based) sample splitting \citep{athey2016recursive,wager2018estimation}. To construct prediction intervals for CATE estimation, we apply AutoCP to construct prediction intervals for $Y(0)$ and for $Y(1)$ separately.  If these prediction intervals are $[a_0, b_0]$ and $[a_1,b_1]$ respectively, then the prediction interval for CATE estimation is $[a_1-b_0, b_1 - a_0]$ and the length of the prediction interval is $(b_1 - a_1) + (b_0 - a_0)$. To be consistent with our previous experiment, we use  95\% prediction intervals (miscoverage $\alpha = 0.05$) for $Y(0), Y(1)$ so that we obtain $95\% \times 95\% = 90.25\% \approx 90\%$ prediction intervals for the response difference $Y(1)-Y(0) | X=X$. We compare the prediction intervals produced by AutoCP (using CMGP as the prediction model and optimizing only the estimators and calibration methods) with the prediction intervals produced by Causal Random Forest (CRF) \citep{wager2018estimation} and the credible sets produced by Causal Multitask Gaussian Process (CMGP) \citep{alaa2017bayesian} and Bayesian Additive Regression Trees (BART), using the datasets IHDP \citep{hill2011bayesian} and LBIDD \citep{shimoni2018benchmarking, mathews1998infant}.\footnote{We omit comparison against estimation methods for CATE that do not produce prediction intervals.}  The results are reported in Table \ref{tab:treatment}. As can be seen, CMGP and AutoCP achieve the $90\%$ target coverage rate for both Response Coverage  and CATE Coverage, but AutoCP provides much narrower prediction intervals. The credible set of CMGP over-covers the data at a price of its interval length. In contrast, both BART and CRF fall short on Response Coverage on both datasets, and BART also falls short on CATE Coverage.

\begin{table}[h]
\centering
\begin{small}
\caption{CATE Estimation on IHDP and LBIDD datasets.  Interval lengths are shown in {\bf bold} if the $90\%$ target coverage rate is satisfied by the intervals for both the Response difference and CATE.}
\label{tab:treatment}
\setlength\tabcolsep{3.2pt}
\begin{adjustbox}{width=1\columnwidth,center}
\begin{tabular}{c|c|c|c}	
\toprule
 \multicolumn{4}{c}{IHDP} \\
\midrule
 \multirow{2}{*}{ Model} & Response &CATE & \multirow{2}{*}{Avg. Length} \\
 & Avg. Coverage ($\%$) &  Avg. Coverage ($\%$) & \\
\midrule
 AutoCP	  & $95.71\pm 1.96$ &  $97.82\pm 2.71$  &$ \bm{ 10.18 \pm2.54}$ \\
 \midrule
CMGP  &  $ 97.79\pm 2.72$ &  $99.95 \pm  0.23$ &$ \bm{15.68 \pm 3.53}$ \\
\midrule
BART  & $33.06\pm 8.23$ & $41.53\pm 13.74$  & $6.13\pm 11.18$ \\
\midrule
CRF   &  $55.61\pm 22.74$ &  $93.94 \pm 3.22$ & $14.89 \pm 19.73$ \\
\midrule
 \multicolumn{4}{c}{LBIDD} \\
\midrule
AutoCP & $95.03\pm  0.68$ &  $100.00\pm  0.00$ & $\bm{3.86 \pm 0.06}$ \\
 \midrule
CMGP  & $99.94\pm 0.08$ &  $100.00\pm  0.00$   & $\bm{6.91\pm 0.14}$ \\
 \midrule
BART  & $8.87\pm 5.79$   & $63.74\pm 33.74$  & $0.15 \pm 0.07$  \\
\midrule
CRF   & $73.67\pm 26.74$   &$100.00\pm 0.00$  & $2.29\pm 0.56$ \\
\bottomrule
\end{tabular}
\end{adjustbox}
\end{small}
\end{table}

Appendix {\color{red}C} provides additional results of our experiments.  We first provide a proof of the concept \quotes{Discrimination} by comparing the length histogram of the intervals constructed by AutoCP and CQR-NN. Then for the first set of experiments, Tables 6 and 7 in the Appendix show the averaged coverage rate and interval length of each algorithm on each of the eight datasets, at target coverage rate 90\% and 95\% respectively.

\section{Conclusion}

This paper introduces AutoCP, which is a simple and powerful AutoML framework for constructing prediction intervals with the user-specified coverage guarantee. Experiments using a variety of datasets demonstrate that AutoCP is sample-efficient in finding the optimal pipeline (Appendix {\color{red}A.3}); AutoNCP outperforms benchmark algorithms (\Cref{sect:experiments}). AutoCP is effective not only for the supervised learning problems  but also for CATE estimation that plays an important role in personalized medicine and policymaking.  Because AutoCP provides tight and valid prediction intervals in real-world applications, it allows non-experts to use prediction models with actionable uncertainty estimates.

\clearpage
\balance
\bibliographystyle{unsrtnat}
\bibliography{bio_aistats}

\end{document}